\documentclass{article}

 \usepackage[main, final]{neurips_2025}
\workshoptitle{NeurIPS 2025 Workshop LXAI}

\usepackage[utf8]{inputenc} 
\usepackage[T1]{fontenc}    
\usepackage{hyperref}       
\usepackage{url}            
\usepackage{booktabs}       
\usepackage{amsfonts}       
\usepackage{nicefrac}       
\usepackage{microtype}      
\usepackage{xcolor}         
\usepackage{graphicx}
\usepackage{threeparttable}
\usepackage{multirow}

\title{Exploration of Incremental Synthetic Non-Morphed Images for Single Morphing Attack Detection}

\author{%
  David Benavente-Rios\thanks{Corresponding author} \\
  Dept. Ing. Industrial\\
  Universidad de Santiago \\
  9170022 Estacion Central, Chile \\
  \texttt{david.benavente.r@usach.cl} \\
  \And
  Juan Ruiz Rodriguez\\
  LAETEC - UPLA \\
  Universidad de Playa Ancha\\
  Valparaíso, Chile \\
  \texttt{juanpablo.ruiz.r@gmail.com} \\
  \And
  Gustavo Gatica \\
  Facultad de Ingeniería\\
  Universidad Andres Bello\\
  Santiago 7500971, Chile \\
  \texttt{ggatica@unab.cl} \\
}

\begin{document}

\maketitle

\begin{abstract}
  This paper investigates the use of synthetic face data to enhance Single-Morphing Attack Detection (S-MAD), addressing the limitations of availability of large-scale datasets of bona fide images due to privacy concerns. Various morphing tools and cross-dataset evaluation schemes were utilized to conduct this study. An incremental testing protocol was implemented to assess the generalization capabilities as more and more synthetic images were added. The results of the experiments show that generalization can be improved by carefully incorporating a controlled number of synthetic images into existing datasets or by gradually adding bona fide images during training. However, indiscriminate use of synthetic data can lead to suboptimal performance. Evenmore, the use of only synthetic data (morphed and non-morphed images) achieves the highest Equal Error Rate (EER), which means in operational scenarios the best option is not relying only on synthetic data for S-MAD.
\end{abstract}

\section{Introduction}
\label{sec1}
In the last few years, the use of fraudulent identification documents has increased, capturing the attention of law enforcement agencies and governments \cite{Ibsen-book}. Because of this, authorities and private institutions have adopted the use of biometric systems to verify the identity of individuals in many contexts (e.g. mobile applications, border checks and airports) using their behavioral and biological characteristics. Nevertheless, these systems may be subject to attacks (e.g. physical or digital) that could lead to vulnerabilities \cite{Ibsen-book}.

One type of digital attack is morphing, defined as the use of morphed images generated by combining face images of two or more subjects into one image. In that sense, to attenuate this risk, many works have been proposed for the automated face Morphing Attack Detection (MAD) \cite{Tapia-FeatureVisualisation-SMAD-IWBF-2023, Dargaud-PCABasedSingleMorphingAttackDetection-WACVw-2023}. The morphing attacks can be classified into two main scenarios: Single-image-based Morphing Attack Detection (S-MAD) and Differential Morphing Attack Detection (D-MAD).

The focus of this work is on S-MAD using features extracted from face recognition systems and "non-morphed"\footnote{Non-morphed images are defined as synthetic images used to create new morphed images.} images from a synthetic dataset. The "non-morphed" are synthetic face images used as bona fide. The main contribution of this proposal is:

\begin{itemize}

    \item Evaluate the impact of integrating "non-morphed" images, which are synthetically generated and shall simulate natural conditions, into the training process to enhance the number of existing bona fide images. This approach is motivated by the limited availability of bona fide images due to privacy concerns related to deep learning methods trained for S-MAD. It is important to note all bona fide images are compliant with ICAO quality requirements \cite{ICAO-9303-p9-2021, Merkle-OFIQ-Report-240930}. 

\end{itemize}

The rest of the paper is organized as follows: Section \ref{sec:related} reviews the related works. Section \ref{sec:datasets} details the databases and metrics. Section \ref{sec:method} explains the method proposed. Section \ref{sec:exp_results} explains the experiments and results obtained, and Section \ref{sec:conclusions} states the conclusions of this work.

\section{Related Work}
\label{sec:related}

In recent years, several works have been presented with the goal of developing methods for S-MAD\cite{Raghavendra-DetectingFaceMorphing-CVIP-2018, Tapia-FeatureVisualisation-SMAD-IWBF-2023}, most of them rely on deep learning-based methods, but there are others which still rely on hybrid or texture features \cite{Tapia-AlphaNet-2023}.

Since the creation of CNNs, many different models have been proposed and used in the context of S-MAD. One of the benefits of the use of CNNs is they are data-efficient models compared to their counterparts, such as the Vision Transformers \cite{Dosovitskiy-vit-2021}. This data efficiency produced by the inductive bias present in the CNN models enables it to remain a suitable option for use in tasks where a meaningful data quantity is not available for training purposes.

Huber et al. \cite{Huber_SYNMAD_2022} presented the Competition on Face Morphing Attack Detection Based on Privacy-aware Synthetic Training Data (SYN-MAD). The competitors were encouraged to develop MAD algorithms using only the Synthetic Morphing Attack Detection Development (SMDD) \cite{Damer_SMDD_2022_CVPR} dataset composed by the "non-morphed" subset with 25k, and the morph subset with 15k. Finally, the competitors tested their MAD algorithms with landmark-based and GAN-based morphed images.

Some works have previously analysed the use and impact of synthetic images for MAD. In \cite{Tapia-ImpactSynImgs-2024}, they proposed a siamese CNN-based scheme using MobileNetV2\cite{Sanlder_mobilenetv2_2018}, MobileNetV3\cite{Howard-MobileNetV3-ICCV-2019} and EfficientNet-B0 \cite{Mingxing-EfficienNet-ICML-2019}. They developed a state-of-the-art (SOTA) database comprised of the Facial Recognition Technology (FERET) database, the Face Recognition Grand Challenge (FRGCv2) database, the Face Research London Lab (FRLL) database and the AMSL database. Besides, The SYN-MAD database of synthetic images and the SMDD test set were also used. Experiments were conducted in these scenarios: train SYN-MAD/test SMDD, train SYN-MAD/test SOTA, train SOTA/test SMDD and train Mix/ test SMDD. They demonstrated that the mixed training using digital and synthetic images may improve the performance of MAD, and the use of only synthetic images could lead to worse results.

\begin{figure*}[h]
\scriptsize
\centering
  \begin{tabular}{c}
  \includegraphics[height=1.4cm, width=1.3cm]{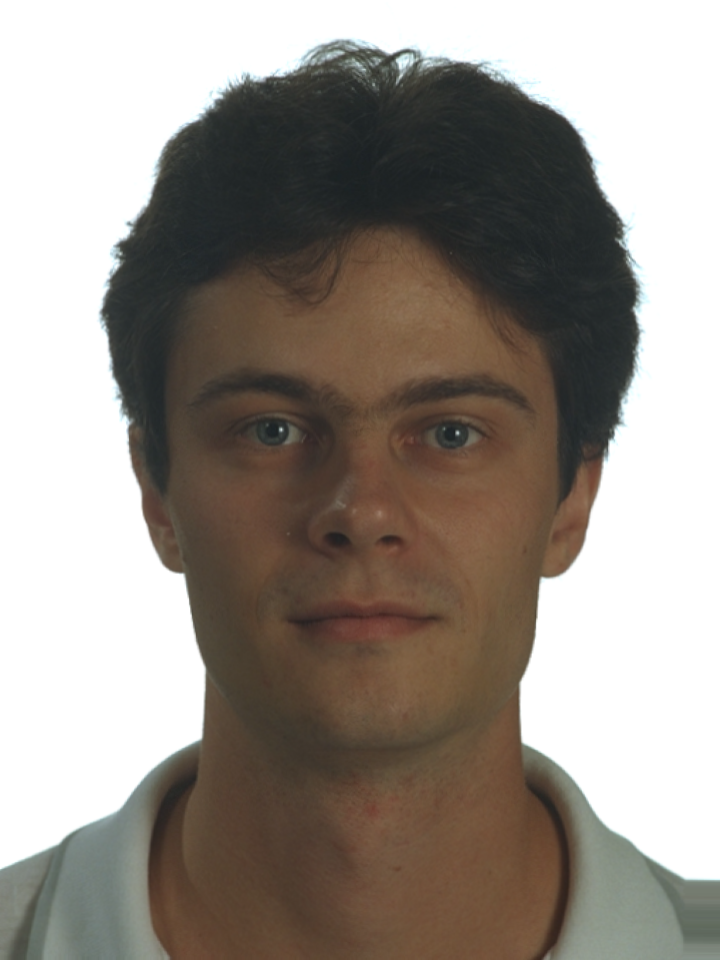}\includegraphics[height=1.5cm, width=1.3cm]{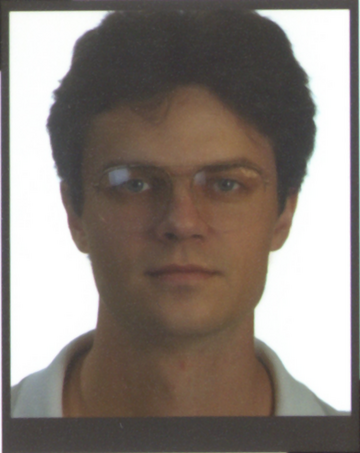}\includegraphics[height=1.4cm, width=1.3cm]{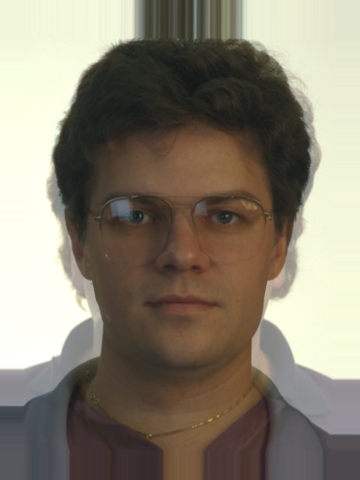}\includegraphics[height=1.4cm, width=1.3cm]{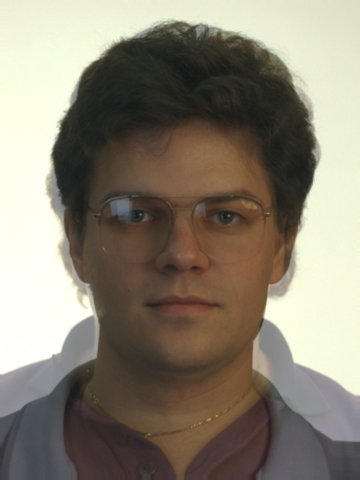}\includegraphics[height=1.4cm, width=1.3cm]{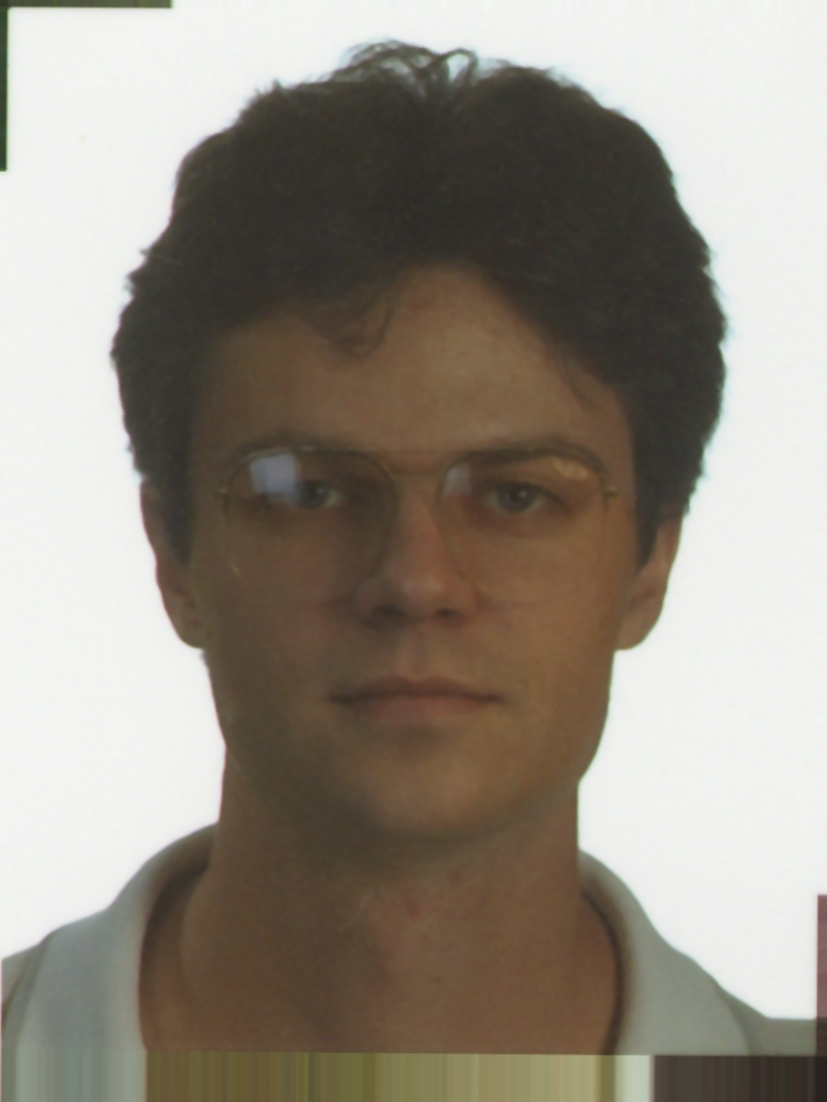}\includegraphics[height=1.4cm, width=1.3cm]{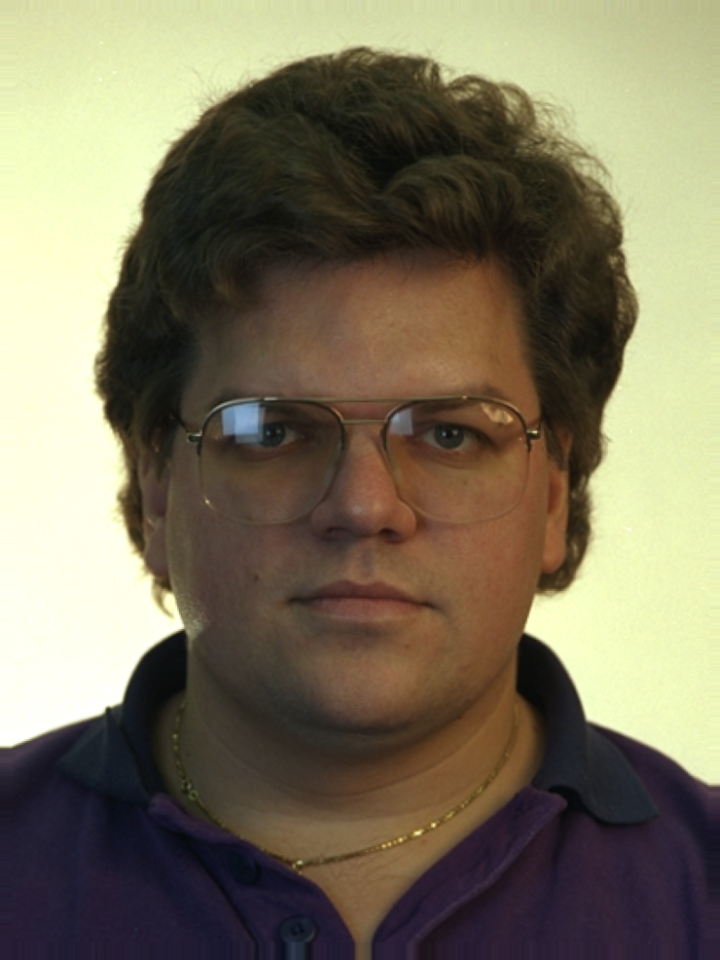}\\
  \includegraphics[height=1.4cm, width=1.3cm]{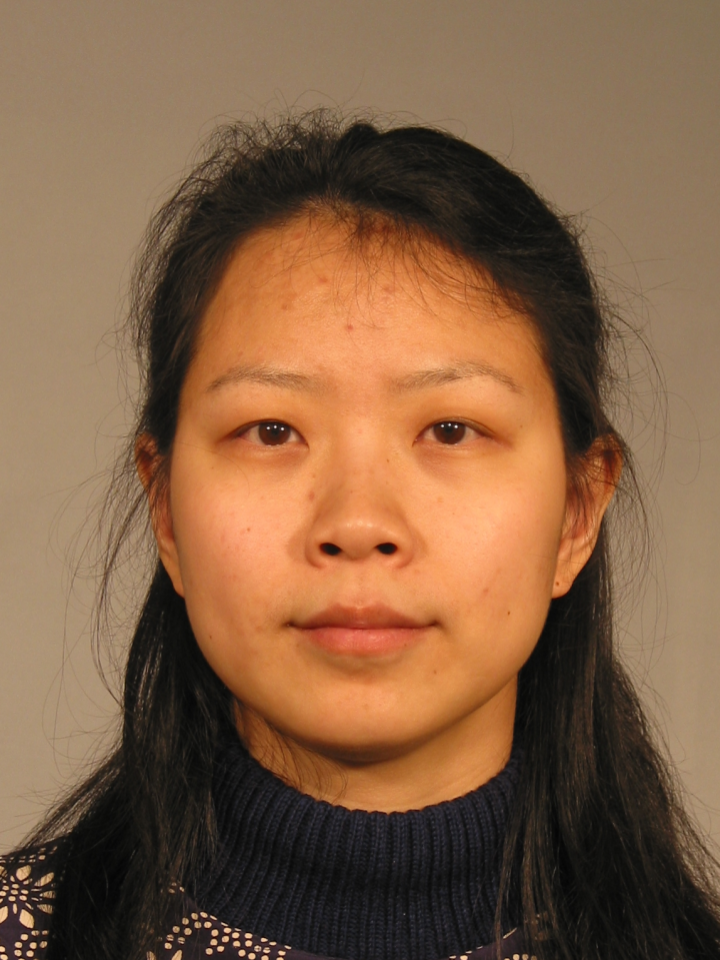}\includegraphics[height=1.5cm, width=1.3cm]{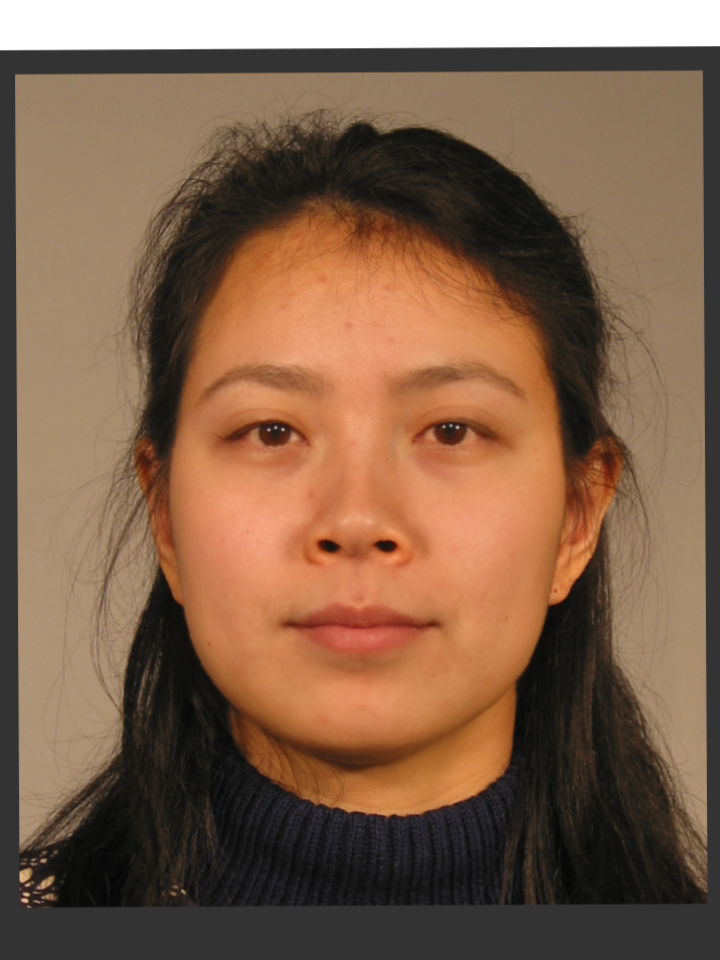}\includegraphics[height=1.4cm, width=1.3cm]{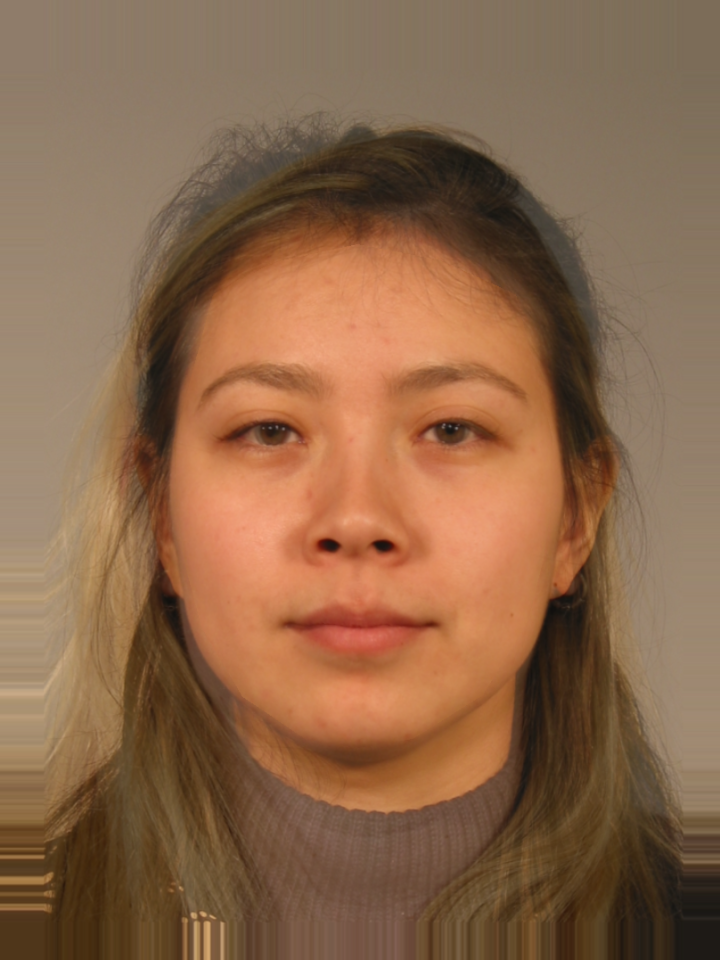}\includegraphics[height=1.4cm, width=1.3cm]{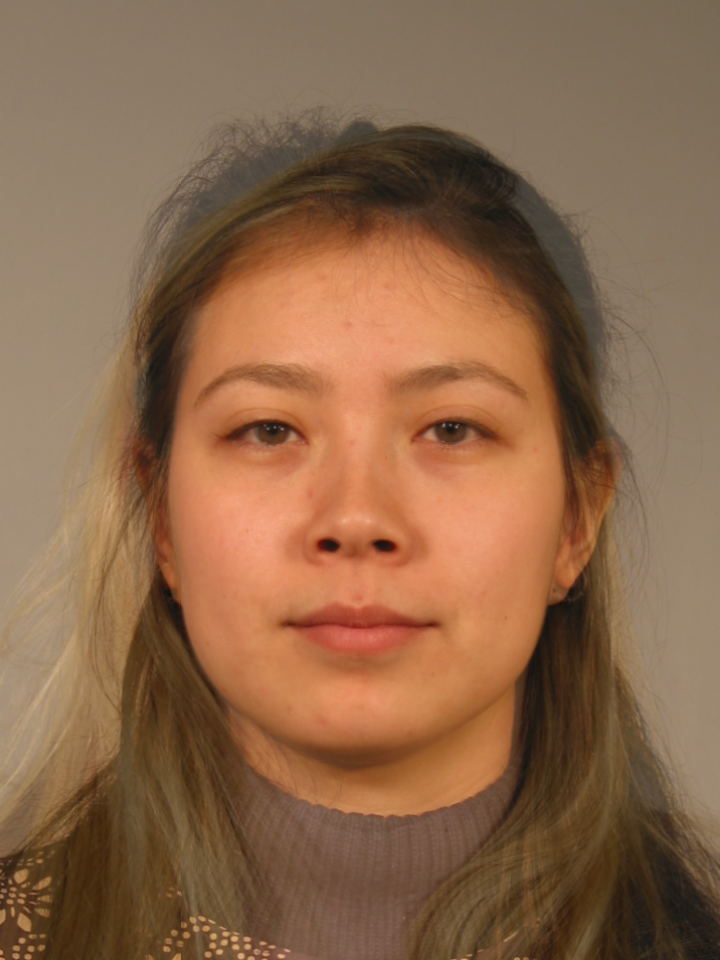}\includegraphics[height=1.4cm, width=1.3cm]{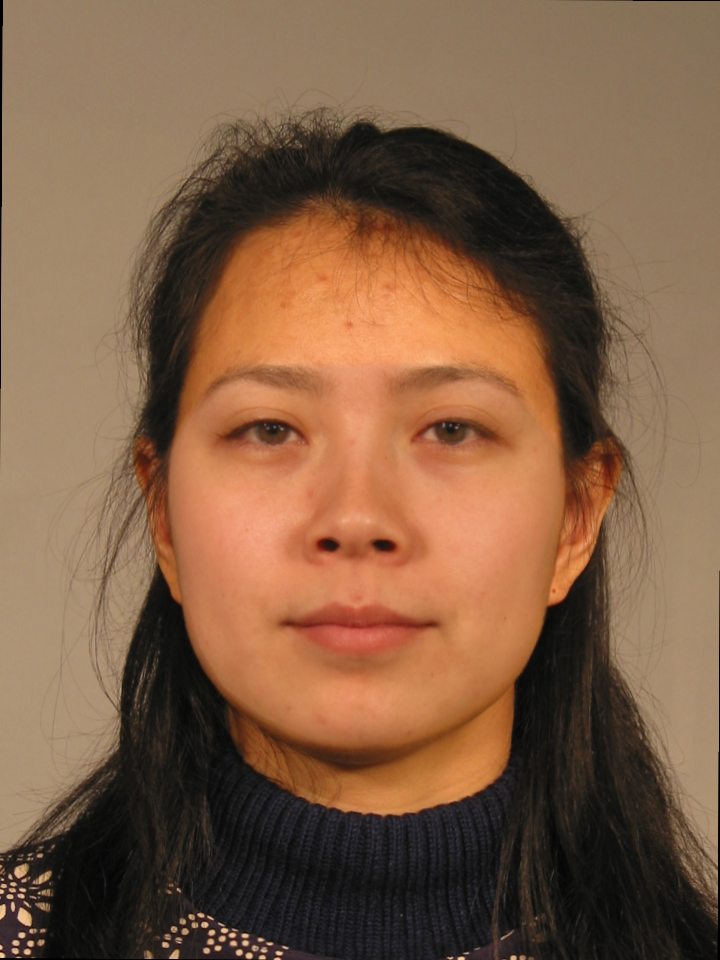}\includegraphics[height=1.4cm, width=1.3cm]{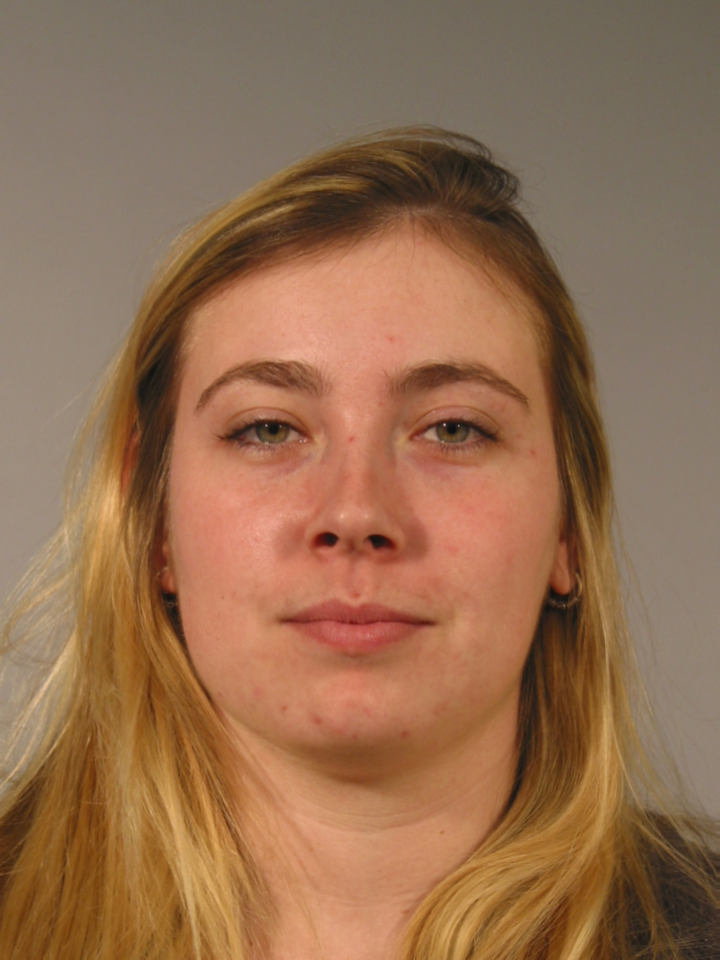}\\
  \includegraphics[height=1.4cm, width=1.3cm]{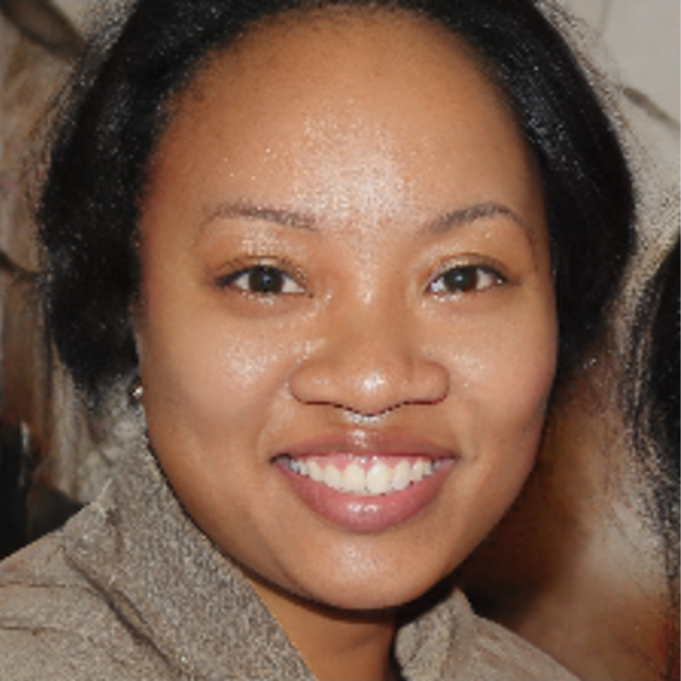}\includegraphics[height=1.4cm, width=1.3cm]{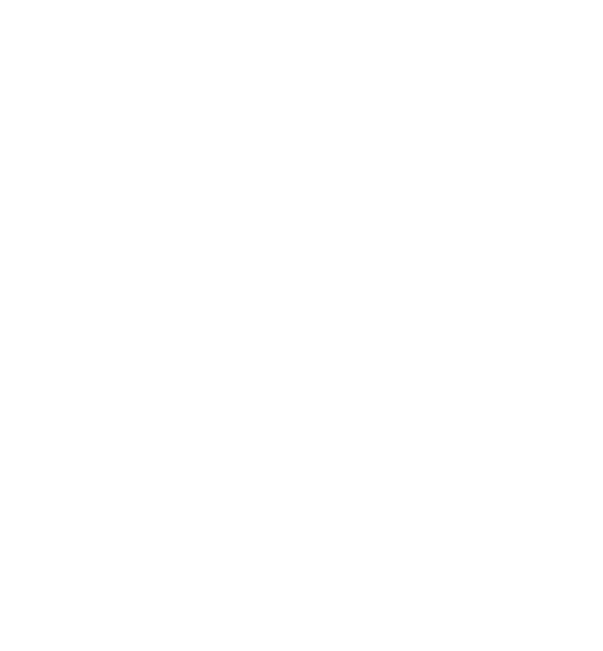}\includegraphics[height=1.4cm, width=1.3cm]{white_space.png}\includegraphics[height=1.4cm, width=1.3cm]{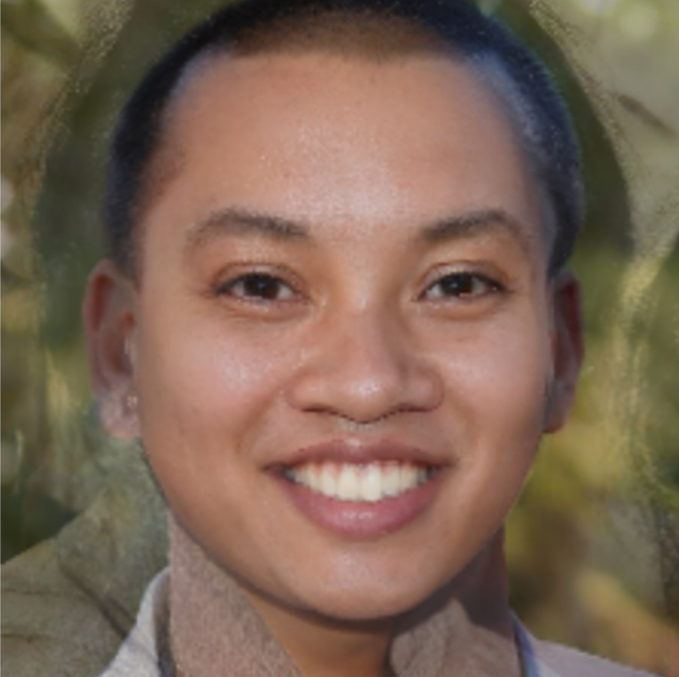}\includegraphics[height=1.4cm, width=1.3cm]{white_space.png}\includegraphics[height=1.4cm, width=1.3cm]{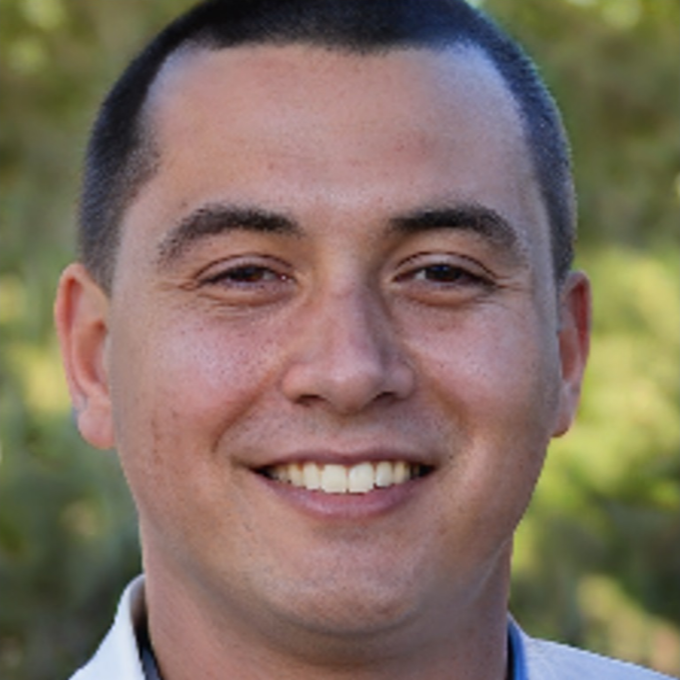}\\
  \end{tabular}
  \caption{Examples of the images from the datasets. In the first row, there are examples from the FERET subset, the second row shows examples from the FRGC subset, and the third row shows the examples of the SMDD dataset. The first column is the bona fide image of the subject 1, the bona fide of the subject 2 is on the last column. Between the subject 1 and subject 2 columns are the morph images: Facefusion \cite{FaceFusion-Morph}, Face morpher\cite{FaceMorpher-Morph}, OpenCV \cite{Bradski-OpenCV-Morph-2000} and UBO \cite{UBO-Morphing}, respectively.}
\label{fig:FERET-FRGV-example}
\end{figure*}

\section{Databases}
\label{sec:datasets}

This work used three datasets: the Facial Recognition Technology database (FERET) \cite{FERET-database-2003} (Figure \ref{fig:FERET-FRGV-example}), the Face Recognition Grand Challenge database (FRGCv2) \cite{Phillips-FRGC-FG2006} (Figure \ref{fig:FERET-FRGV-example}) and the Synthetic Morphing Attack Detection Development (SMDD) dataset (Figure \ref{fig:FERET-FRGV-example}). Examples of these datasets are depicted in Figure \ref{fig:FERET-FRGV-example}.

\subsection{FERET dataset}
The set used is a subset part of the Colour FERET Database provided by the National Institute of Standards and Technology (NIST)
Face Recognition Technology (FERET) program. Different image processing methods were used: print/scan 300 (PS300), print/scan 600 (PS600) and digital resized.
\subsection{FRGC dataset}
The second version of the Face Recognition Grand Challenge (FRGCv2) dataset is used. Furthermore, for this dataset, the same image processing methods as those employed in FERET were used.
\subsection{SMDD dataset}
The Synthetic Morphing Attack Detection Development (SMDD) dataset \cite{Damer_SMDD_2022_CVPR} is a synthetic dataset composed of 40k images, where 15K are attacks and 25K are bona fide samples. The morphed subset is composed of samples generated with Generative Adversarial Network (GAN)\cite{Goodfellow_GAN_2014} StyleGAN2-ADA\cite{Karras_stylegan2_2020}.

\subsection{Morphs}
The morphed image subsets of the FERET and FRGCv2 were generated by four morphing tools, and these are listed below:
\begin{itemize}
    \item FaceFusion \cite{FaceFusion-Morph}: A mobile application which generates realistic faces with morphing artefacts that are almost imperceptible.
    \item Face Morpher \cite{FaceMorpher-Morph}: Open-source implemented in Python and depends on the package STASM. Morphed generated with this tool maintains many visible artefacts.
    \item OpenCV \cite{Bradski-OpenCV-Morph-2000}: This open-source tool uses Dlib for the detection of face landmarks, but as a result, some of the morph artefacts prevail.
    \item UBO \cite{UBO-Morphing}: Morphing tool developed at the University of Bologna (UBO). The tool calculates the facial landmarks of two subjects and combines them to generate the morphed image. 
\end{itemize}

On the other hand, the morphed subset of the SMDD dataset are not specified because these images were generated by a GAN model. Table \ref{tab:datasets-summary} presents the summary of the datasets used.

\begin{table}[t]
\caption{Summary of the dataset used with the processing methods and morph tools.}
\label{tab:datasets-summary}
\centering
\begin{threeparttable}[b]
\begin{tabular}{llllll}
\hline
Dataset & Subjects & Bona-fide & Morph   & Tool                                                                             & Notes\\
\hline
FERET   & 529      & 529*3     & 529*4*3 & \begin{tabular}[l]{@{}l@{}}FaceFusion\\ Face Morpher\\ OpenCV\\ UBO\end{tabular} & \begin{tabular}[l]{@{}l@{}}PS300\tnote{1}\\ PS600\tnote{1} \\ Resized\tnote{1} \end{tabular} \\
\hline
FRGCv2  & 533      & 984*3     & 964*4*3 & \begin{tabular}[l]{@{}l@{}}FaceFusion\\ Face Morpher\\ OpenCV\\ UBO\end{tabular} & \begin{tabular}[l]{@{}l@{}}PS300\tnote{1} \\ PS600\tnote{1} \\ Resized\tnote{1} \end{tabular} \\
\hline
SMDD    & -        & 15,000    & 25,000  & OpenCV                                                                           & \begin{tabular}[l]{@{}l@{}}Synthetic \\ morphed \\ images\end{tabular}\\
\hline
        & Total    & 19,539    & 42,916  & &
\end{tabular}
\begin{tablenotes}
       \item [1] Image processing-methods: print/scan 300dpi (PS300), print/scan 600dpi (PS600) and digital-resized.
     \end{tablenotes}
    \end{threeparttable}
\end{table}

\section{Method}
\label{sec:method}

To develop this deep learning-based method, we used a dataset $D$ conformed by $X$ and $Y$(see eq. \ref{eq:3}):

\begin{equation}
    D = (x_{k}, y_{k}) \mid x_{k} \in \{R^{3 \times H \times W}\}, y_{k} \in \{0, 1\},k =\{ 1, 2, \dots, n\},
    \label{eq:3}
\end{equation}

where  $x_{k} \subset X, y_{k} \subset Y$. Thus, $D$ was used to train the dl-based model $ f_{\theta}(x_{k}; \theta)$ where $f_{\theta}: X \rightarrow Y$, $\theta$ are parameters of the CNN, learned in the training procedure, and $ \hat{y_{i}}$ are the outputs produced by $ f_{\theta}(x_{k}, \theta)$ (see eq. \ref{eq:4}):

\begin{equation}
    \hat{y_{k}} = f_{\theta}(x_{k}; \theta) / \hat{y_{k}} \in \{0, 1\}
    \label{eq:4}
\end{equation}

where $\hat{y_{k}} \subset \hat{Y}$ is the outcome. In this case we have $x_{k} \in R^{3xHxW}$ for the input $X$ which are the RGB images, $y_{k} \in \{0, 1\}$ for the labels $Y$, and $\hat{y_{k}} \in \{0,1\}$ for the outputs $\hat{Y}$.

In this way, the models were trained using $D$ and the Adam optimizer. Besides, we tested the Stochastic Gradient Descent (SGD) approach. The grid search method was applied to find the best learning rate: $1e-3, 1e-4$ and $1e-5$. Furthermore, a batch size of $64$ was employed, the model was trained for $100$ epochs, $n$ is the number of samples per batch and categorical cross-entropy worked as the loss function (eq. \ref{eq:5}).
\begin{equation}
   L_{i} = H(y_{k}, \hat{y_{k}}) = -\frac{1}{n}\sum_{x}^{} (y_{k}) \log (\hat{y_{k}})
   \label{eq:5}
\end{equation}

\begin{figure}[b]
\scriptsize
\centering
              \includegraphics[height=5cm, width=\textwidth]{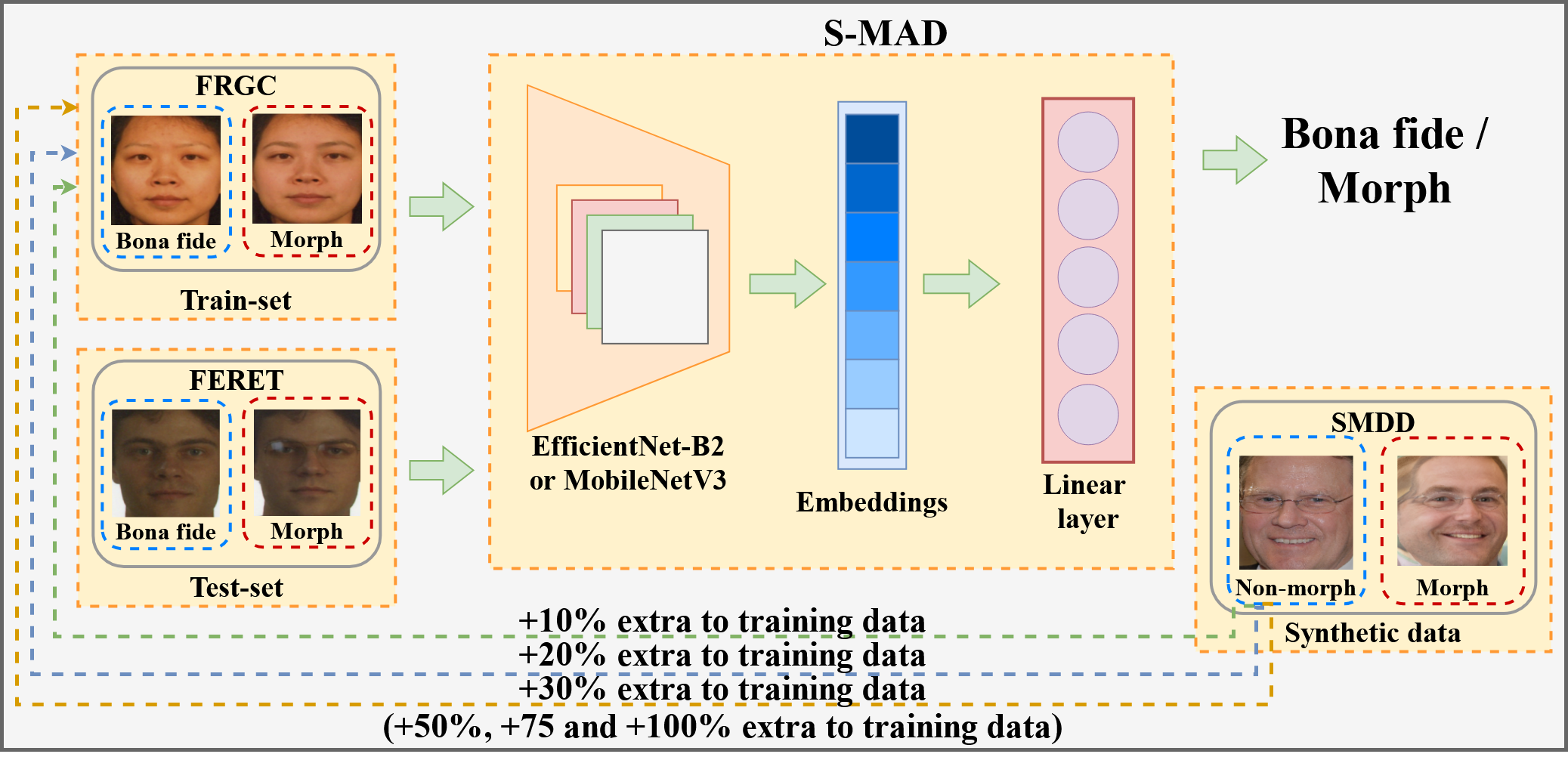}\\
    
    \caption{S-MAD workflow, training with FRGCv2 and testing with FERET.}
    \label{fig:smad-diagram}
\end{figure} 

\subsection{Random Sampling}
Different scenarios were generated to evaluate the effects of the synthetic data, adding non-morphed data from the SMDD to the bona fide subsets of the training set. These experiments are listed below: 

\begin{itemize}
    \item Training without synthetic data (see exp. \ref{exp-1})
    \item Adding a percentage of non-morphed synthetic data (10\%, 20\%, 30\%, 50\%, 75\% and 100\%) (see exp. \ref{exp-2})
    \item Using only synthetic data to the training set (see exp. \ref{exp-3})
\end{itemize}

The synthetic data added from the dataset $D_{SMDD}$ is defined as follows (see eq. \ref{eq:6}):

\begin{equation}
    D_{SMDD} = \{(x_{i}, y_{i}) / x_{i} \in R^{3 \times H \times W}\}, y_{i} \in \{0, 1\}, i =\{ 1,\dots, n\},
    \label{eq:6}
\end{equation}

To add the synthetic data to the training set we took a random sample $S_{(SMDD,j)}$ from $D_{SMDD}$, as follows:

\begin{equation}
    S_{(SMDD,j)} \subset D_{SMDD}
\end{equation}

where $j$ represent the percentage increased to the original dataset, $\big|S_{(SMDD,j)} \big|= m$ and $m < n$. Then $S_{(SMDD,j)}$ was incorporated to the training set $D_{train}$. A diagram that summarizes all the processes is presented in Figure \ref{fig:smad-diagram}).

\subsection{Preprocessing}
\begin{figure}[h]
\scriptsize
\centering
    \begin{tabular}{ccc}
              \includegraphics[height=1.4cm, width=1.3cm]{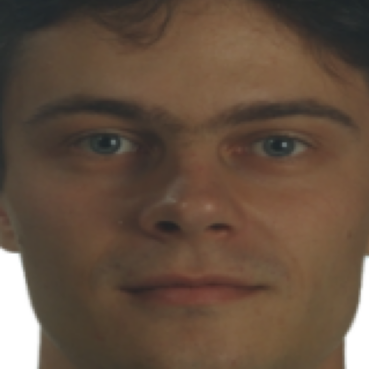}\includegraphics[height=1.4cm, width=1.3cm]{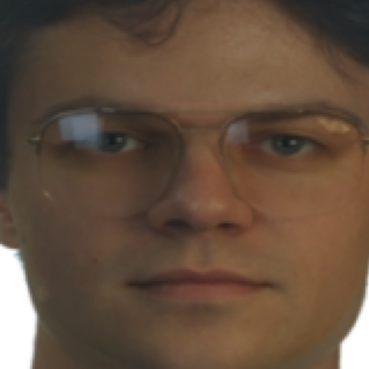}&\includegraphics[height=1.4cm, width=1.3cm]{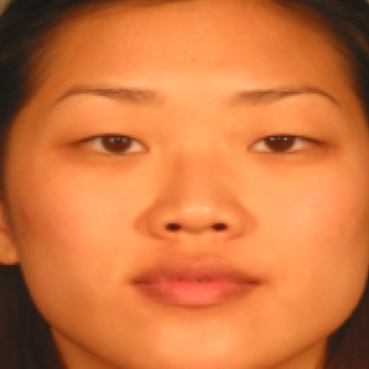}\includegraphics[height=1.4cm, width=1.3cm]{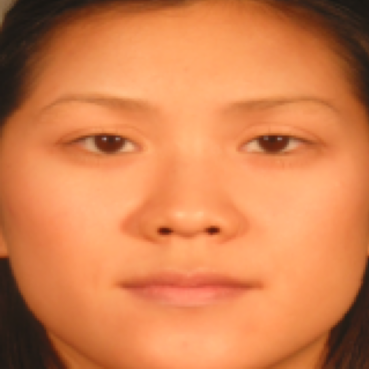}&\includegraphics[height=1.4cm, width=1.3cm]{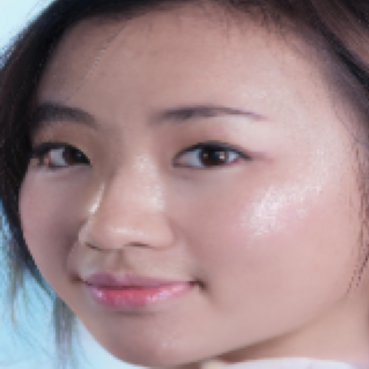}\includegraphics[height=1.4cm, width=1.3cm]{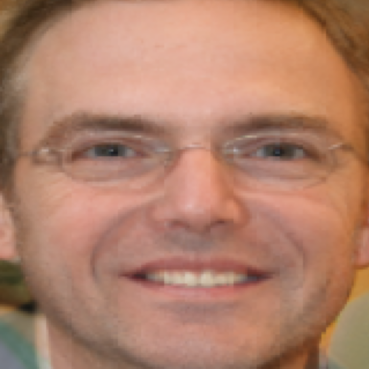}\\
    
    a. & b. & c.\\
    \end{tabular}
    \caption{Examples of the pre-processing step done with MTCNN \cite{MTCNN} to FERET (a.), FRGCv2 (b.) and SMDD (c.) datasets, respectively. Each image shows side-by-side bona fide and morph.}
    \label{fig:preproc-example}
\end{figure}

Some works have studied the effects of the context of images in S-MAD \cite{Pimenta-Context-Morphing-SMAD-BIOSIG-2023}, and motivated for the result, was decided to use the Multi-Task Cascade Convolutional Neural Network tool (MTCNN) \cite{MTCNN} as alignment method with the application of scaling factor equal to $0.9$ which result in an image size of $369\times369$ (see Figure \ref{fig:preproc-example}). Afterwards, these data augmentation steps for the training set were applied: Resizing to $224\times224$, random horizontal flip, random rotation, colour jitter and normalization to the stats of the ImageNet-1K dataset.

\subsection{Models}
For our experiments, we employ EfficientNet and MobileNet as backbone architectures. These models are designed to achieve a favorable balance between accuracy and computational efficiency through compound scaling and depthwise separable convolutions. Their reduced parameter count and FLOP requirements make them particularly suitable for scenarios such as border control and mobile identity verification, where real-time inference and limited hardware resources are critical constraints. By adopting these efficient architectures, we ensure that our evaluation framework remains reproducible and aligned with realistic deployment conditions, while avoiding the unnecessary computational overhead of excessively large models.

\subsubsection{MobileNetV3-large}
As a result of the network architecture search (NAS) through the model design and the automated search algorithms, this network emerges as a response for the operability in the context of mobile phone CPUs. The two versions of this MobileNetV3, small and large, offer solutions for the operability in low and high-resource devices \cite{Howard-MobileNetV3-ICCV-2019}. For this work, the MobileNetV3-large was used, which has around 7.7 million trainable parameters.
\subsubsection{EfficientNet-B2} 
The family of EfficientNets were created as a result of the search for the best accuracy with model scaling subject to the combinations of depth, resolution and width based on the proposed compound coefficient \cite{Mingxing-EfficienNet-ICML-2019}. EfficientNet-B2 was used for this work, which has 2.9M. trainable parameters.

\subsection{Metrics}
The metrics used for the evaluation of this method are Morphing Attack Classification Error Rate (MACER) and Bona fide Presentation Classification Error Rate (BPCER). These metrics are established by the International Organization for Standardization (ISO) and are defined in the ISO/IEC DIS 20059 \cite{ISO-IEC-20059}.

\subsubsection{Morphing Attack Classification Error Rate (MACER)}
Is the proportion of morphed samples incorrectly classified as bona fide sample (no morph) in a specific scenario, and is calculated as follows:
\begin{equation}
    MACER = \frac{1}{N_{M}} \sum_{i=1}^{N_{M}} (1 - RES_{i})
\end{equation}

where $\displaystyle N_{M}$ is defined as the number of morphed samples and the value of $\displaystyle RES_{i}$ is $0$ or $1$ if the system classifies to the $\displaystyle i$-th sample as no morph and morph, respectively.

\subsubsection{Bona fide Presentation Classification Error Rate (BPCER)}
Is the percentage of bona fide samples which are wrongly classified by the system as morphed samples, and is calculated as follows:

\begin{equation}
    BPCER = \frac{\sum_{i=1}^{N_{BF}} RES_{i}}{N_{BF}}
\end{equation}
Where $\displaystyle N_{BF}$ is the number of bona fide samples.

Furthermore, based on these metrics, we report: i) Detection Error Trade-off ($DET$) between $MACER$ and $BPCER$; ii) $BPCER$s at different MACER values and iii) the Detection Equal Error Rate ($D-EER$), which is the point where MACER and BPCER are equal.

\section{Experiments and Results}
\label{sec:exp_results}
Two rounds of experiments were executed using FERET as a training set and FRGCv2 as a test set in a cross-dataset evaluation scheme, and vice versa. We developed a subset of experiments from these rounds by adding the synthetic data from SMDD. The dataset was split into $80/20$ was applied to training set to get a validation set. After the data augmentation process was employed, Adam as an optimizer for the training process was set with $\beta1=.99$ and $\beta2=.999$, and the loss function used was categorical cross-entropy. From the configurations of the learning rate tested $1e-5$ achieved the best results. The weights of ImageNet-1K were set as a starting point for the CNNs' training process.

Three different experiment scenarios were developed to study the effects of the synthetic data in the training process for S-MAD (\ref{exp-1}, \ref{exp-2} and \ref{exp-3}). Additionally, all of them were executed using an NVIDIA-A100-80GB GPU from an internal cluster. The experiments configured to run with $8$ worker. Furthermore, the total duration of the experimental process was approximately seven days.

\subsection{Experiment 1 - Without adding synthetic data}
\label{exp-1}
In this experiment, the networks were trained without the addition of synthetic data, which means FERET is used for training, FRGCv2 is used for testing, and vice versa.

\subsection{Experiment 2 - Adding synthetic data}
\label{exp-2}
In this experiment, the networks were trained by adding a random samples from the non-morphed subset of the SMDD to the training set in the incremental proportion of 10\%, 20\%, 30\%, 50\%, 75\% and 100\%. Tables \ref{tab:resutls-summary1} and \ref{tab:resutls-summary2} shows the detail of the samples added in the different rounds of experiments, using FERET and FRGCv2 for train and test, and vice versa.

\subsection{Experiment 3 - Training only with synthetic data}
\label{exp-3}
For this experiment, the SMDD dataset was used as the training set. In other words, the non-morphed and morph subset were employed for training both networks (EfficientNet-B2 and MobileNetV3-large). Finally, the test results were computed over the FERET, and FRGCv2.

\subsection{Results}
\begin{figure*}[t]
\scriptsize
\centering
\includegraphics[height=4cm, width=3cm]{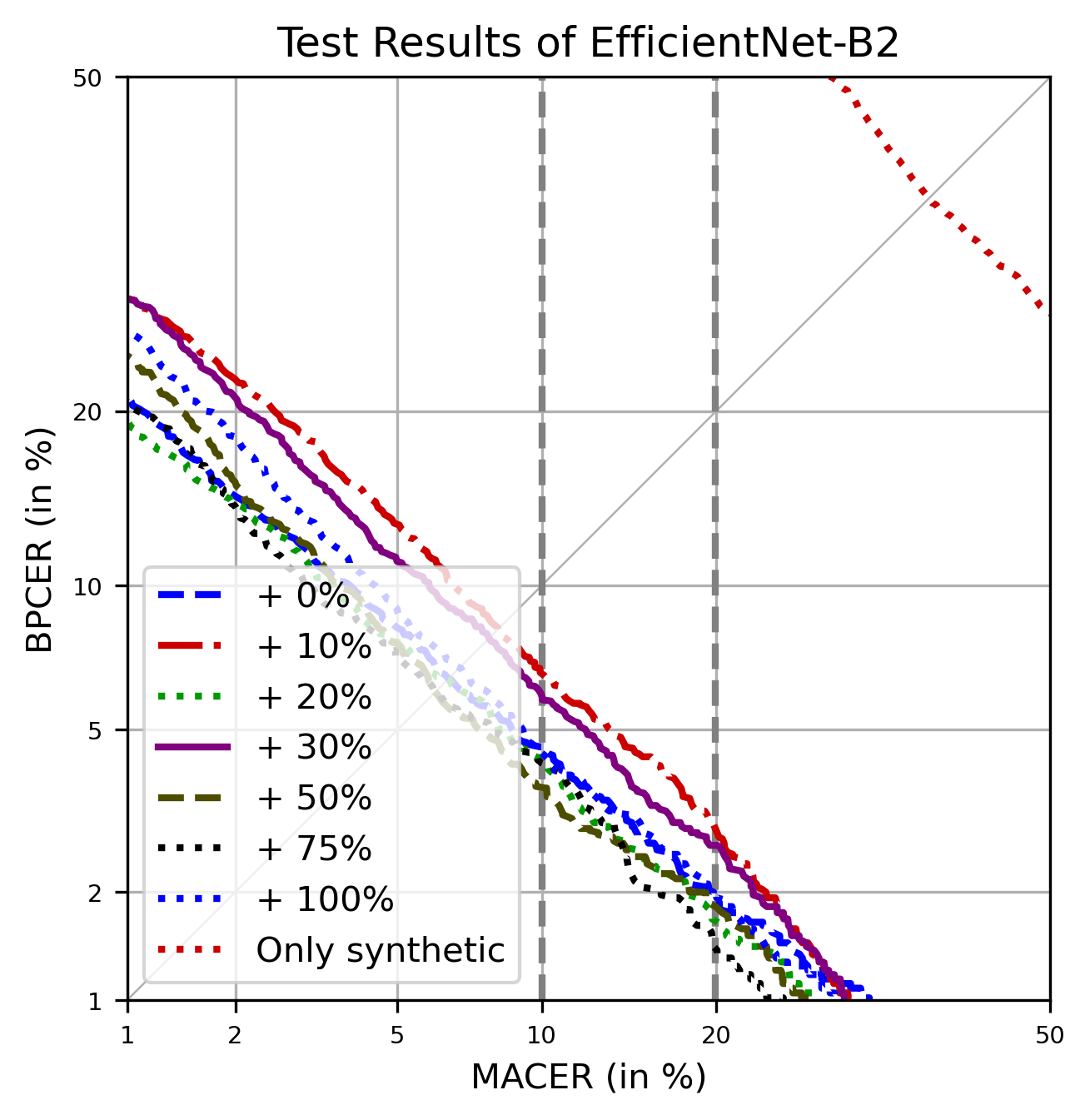}
\includegraphics[height=4cm, width=3cm]{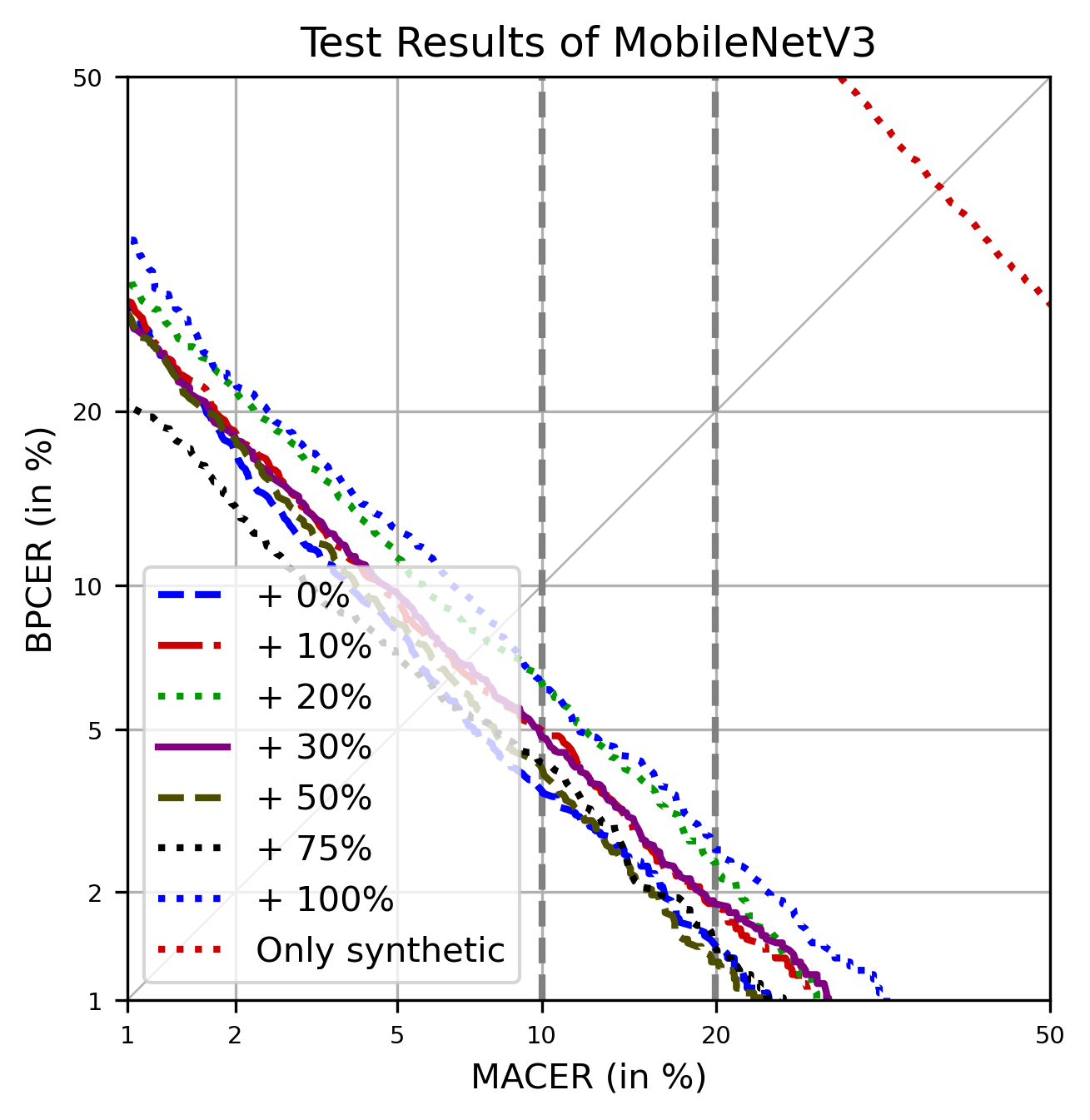}
\includegraphics[height=4cm, width=3cm]{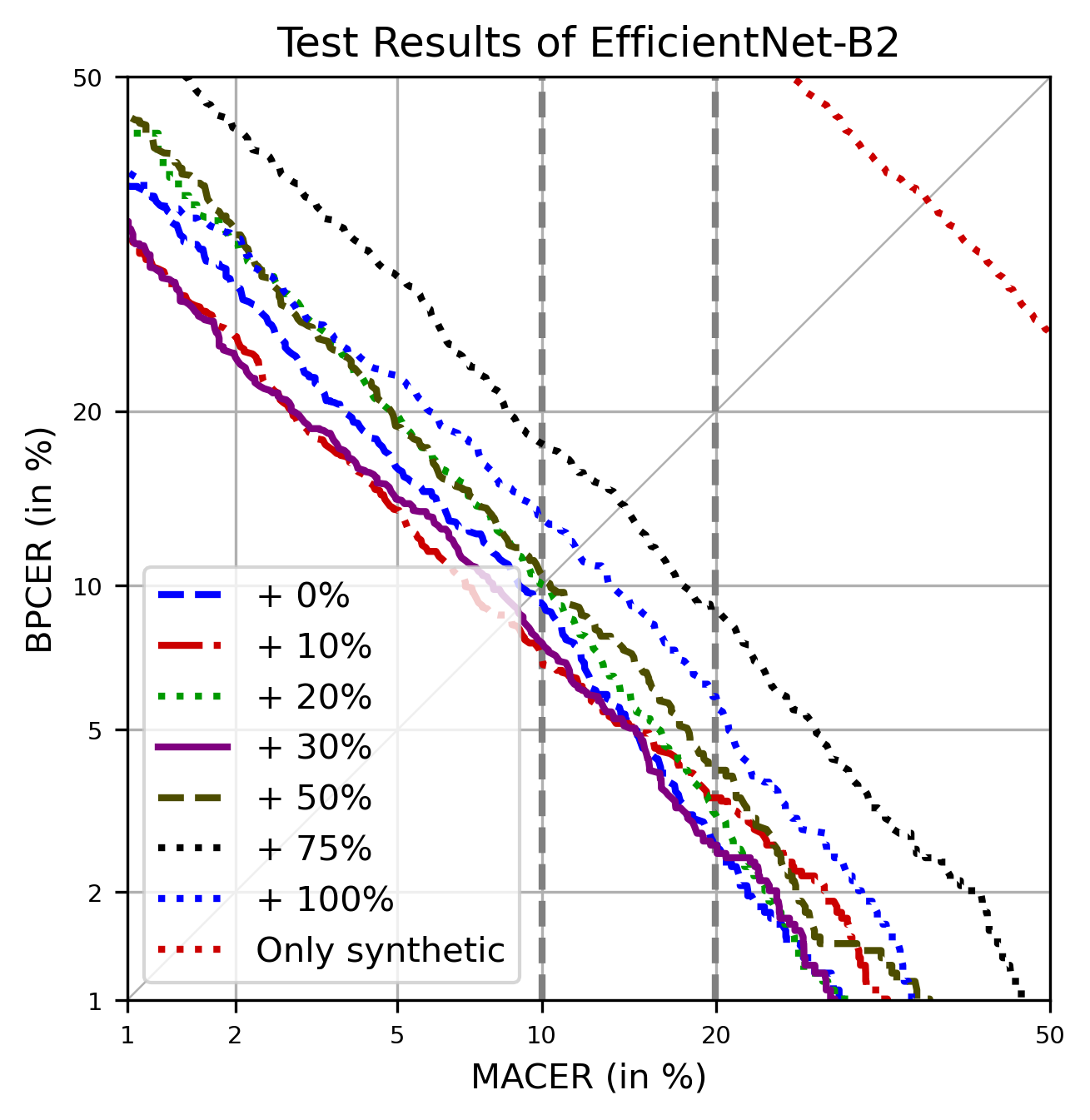}
\includegraphics[height=4cm, width=3cm]{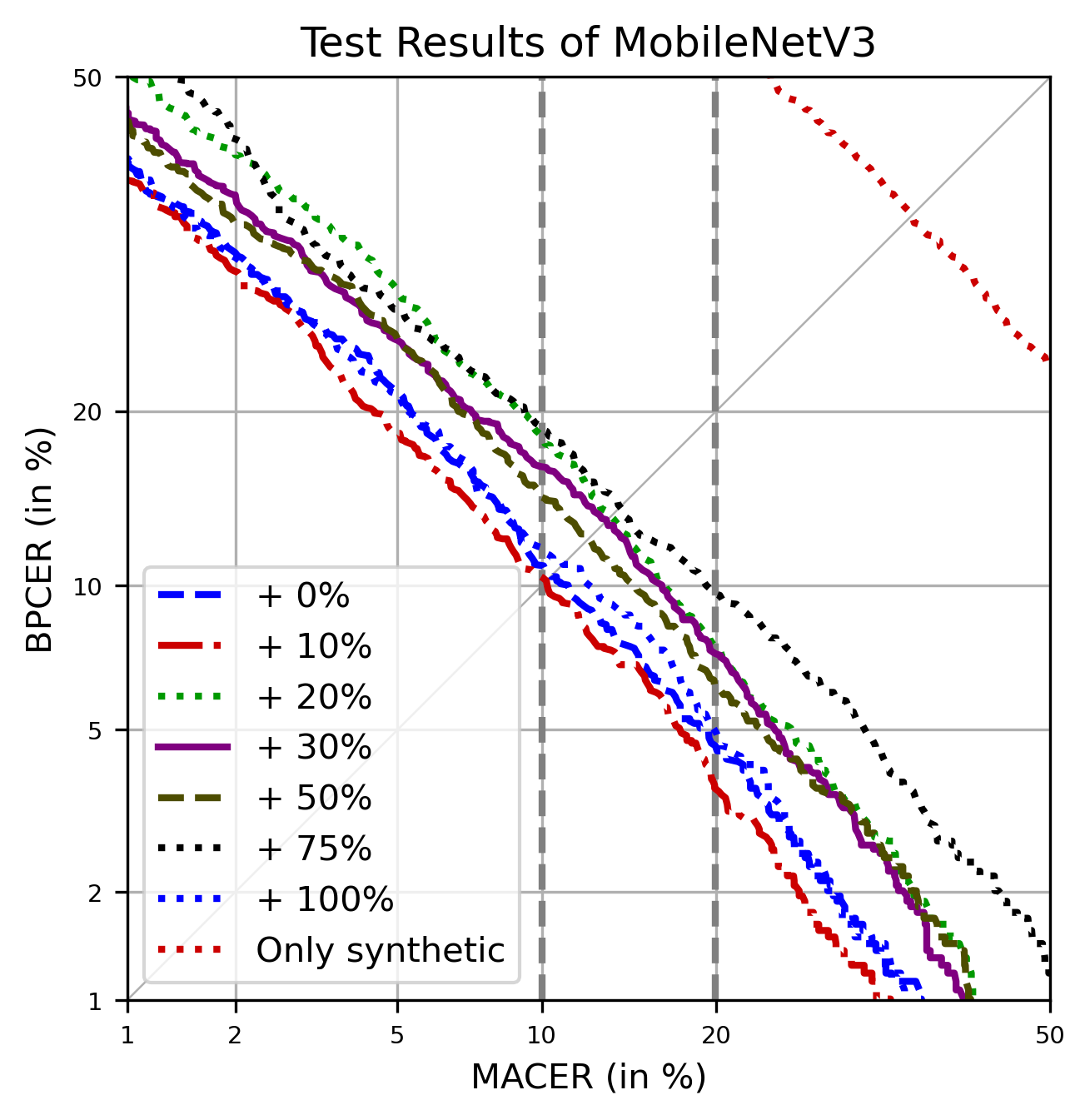} \\
\caption{DET curves of the models achieved with the test set, first and second columns FRGCv2 was employed as test dataset. The third and fourth columns are the results employing FERET as a test dataset.}
    \label{fig:DETs-plots}
\end{figure*}

The outcomes of all experiments conducted are presented in Tables \ref{tab:resutls-summary1} and \ref{tab:resutls-summary2} including the $EER$ and $BPCER$s (5, 10 and 20) operational points. Additionally, the DET curves plotted are presented in Figure \ref{fig:DETs-plots}. 

\subsubsection{Training with FERET and testing with FRGC}
The best performance was achieved by EfficientNet-B2 with the lowest $EER$ equal to $6.09\%$; on the other hand, with the MobileNetV3-large, the lowest $EER$ achieved was $6.10\%$. Both of these results were obtained by adding $75\%$ of synthetic non-morphed data to the training process. 

As is shown in Figure \ref{fig:DETs-plots}, the higher $EER$ with both models was obtained training only with synthetic data, morphed and non-morphed (red line). Overall, with EfficientNet-B2 the mean $EER$ ($\overline{EER}$) obtained is $6.86\%$ and with MobileNetV3-large is $\overline{EER}$ of $6.93\%$.

\begin{table}[b]
\centering
\scriptsize
\begin{threeparttable}[b]
\caption{Summary of results obtained training with FERET and testing with FRGCv2. The best results are in bold.}
\label{tab:resutls-summary1}
\begin{tabular}{llllllll}
\hline
\multirow{2}{*}{Models}  & \multirow{2}{*}{\begin{tabular}[c]{@{}l@{}}Additional\\ data (\%)\end{tabular}} & \multirow{2}{*}{\begin{tabular}[c]{@{}l@{}}Size of the\\ Sample\end{tabular}} & \multirow{2}{*}{\begin{tabular}[c]{@{}l@{}}Total\\ Bona fide\end{tabular}} & \multicolumn{4}{c}{Metrics (\%)}                              \\ \cline{5-8} 
                         &                                                                                 &                                                                               &                                                                            & D-EER         & BPCER5        & BPCER10       & BPCER20       \\ \hline
EfficientNet-B2          & \multirow{2}{*}{0}                                                              & \multirow{2}{*}{-}                                                            & \multirow{2}{*}{1,587}                                                     & 6.47          & 1.97          & 4.57          & 8.27          \\ \cline{1-1} \cline{5-8} 
MobileNetV3              &                                                                                 &                                                                               &                                                                            & 6.17          & 1.42          & 3.59          & 8.23          \\ \hline
EfficientNet-B2          & \multirow{2}{*}{10\%}                                                           & \multirow{2}{*}{160}                                                          & \multirow{2}{*}{1,747}                                                     & 8.20          & 2.85          & 6.67          & 13.01         \\ \cline{1-1} \cline{5-8} 
MobileNetV3              &                                                                                 &                                                                               &                                                                            & 6.80          & 1.86          & 4.98          & 9.28          \\ \hline
EfficientNet-B2          & \multirow{2}{*}{20\%}                                                           & \multirow{2}{*}{320}                                                          & \multirow{2}{*}{1,907}                                                     & 6.47          & 1.69          & 4.23          & 7.69          \\ \cline{1-1} \cline{5-8} 
MobileNetV3              &                                                                                 &                                                                               &                                                                            & 7.77          & 2.30          & 6.30          & 11.35         \\ \hline
EfficientNet-B2          & \multirow{2}{*}{30\%}                                                           & \multirow{2}{*}{480}                                                          & \multirow{2}{*}{2,067}                                                     & 7.86          & 2.64          & 5.93          & 11.11         \\ \cline{1-1} \cline{5-8} 
MobileNetV3              &                                                                                 &                                                                               &                                                                            & 6.94          & 1.86          & 4.81          & 9.62          \\ \hline
EfficientNet-B2          & \multirow{2}{*}{50\%}                                                           & \multirow{2}{*}{800}                                                          & \multirow{2}{*}{2,387}                                                     & 6.09          & 1.83          & 3.70          & 7.66          \\ \cline{1-1} \cline{5-8} 
MobileNetV3              &                                                                                 &                                                                               &                                                                            & 6.53          & 1.29          & 4.07          & 8.44          \\ \hline
\textbf{EfficientNet-B2} & \multirow{2}{*}{\textbf{75\%}}                                                  & \multirow{2}{*}{\textbf{1,200}}                                               & \multirow{2}{*}{\textbf{2,787}}                                            & \textbf{6.09} & \textbf{1.39} & \textbf{4.20} & \textbf{7.08} \\ \cline{1-1} \cline{5-8} 
\textbf{MobileNetV3}     &                                                                                 &                                                                               &                                                                            & \textbf{6.10} & \textbf{1.05} & \textbf{3.05} & \textbf{7.62} \\ \hline
EfficientNet-B2          & \multirow{2}{*}{100\%}                                                          & \multirow{2}{*}{1,587}                                                        & \multirow{2}{*}{3,174}                                                     & 6.81          & 1.97          & 4.40          & 8.91          \\ \cline{1-1} \cline{5-8} 
MobileNetV3              &                                                                                 &                                                                               &                                                                            & 8.17          & 2.58          & 6.23          & 12.74         \\ \hline
EfficientNet-B2          & \multirow{2}{*}{\begin{tabular}[c]{@{}l@{}}Only\\ synthetic\end{tabular}}       & \multirow{2}{*}{-}                                                            & \multirow{2}{*}{25,000}                                                    & 37.96         & 61.24         & 76.51         & 85.71         \\ \cline{1-1} \cline{5-8} 
MobileNetV3              &                                                                                 &                                                                               &                                                                            & 38.95         & 62.32         & 73.73         & 83.80         \\ \hline
\end{tabular}
    \end{threeparttable}
\end{table}

\subsubsection{Training with FRGC and testing with FERET}
The outcomes obtained from all experiments conducted using FRGCv2 as a training set and the FERET version of training and test sets are in the table \ref{tab:resutls-summary2}. 

As is it shown in the Figure \ref{fig:DETs-plots}, something similar to the inverse scheme happened, the higher $EER$ with both models was obtained training only with synthetic morphed and non-morphed data. 

The best performance was achieved with the EfficientNet-B2 with the lowest $EER$ was $8.68\%$; on the other hand, with the MobileNetV3-large, the lowest $EER$ achieved was $10.20\%$. Both these results were obtained by adding only a $10\%$ of synthetic data to the training process. 

Overall, $\overline{EER}$ obtained with EfficientNet-B2 is $10.46\%$ and MobileNetV3-large with an $\overline{EER}$ of $12.05\%$.

\begin{table}[t]
\centering
\scriptsize
\begin{threeparttable}[b]
\caption{Summary of results obtained training with FRGCv2 and testing with FERET. The best results are highlighted in bold.}
\label{tab:resutls-summary2}
\begin{tabular}{llllllll}
\hline
\multirow{2}{*}{Models}  & \multirow{2}{*}{\begin{tabular}[c]{@{}l@{}}Additional\\ data (\%)\end{tabular}} & \multirow{2}{*}{\begin{tabular}[c]{@{}l@{}}Size of \\ the Sample\end{tabular}} & \multirow{2}{*}{\begin{tabular}[c]{@{}l@{}}Total\\ Bona fide\end{tabular}} & \multicolumn{4}{c}{Metrics (\%)}                                 \\ \cline{5-8} 
                         &                                                                                 &                                                                                &                                                                            & D-EER          & BPCER5        & BPCER10        & BPCER20        \\ \hline
EfficientNet-B2          & \multirow{2}{*}{0}                                                              & \multirow{2}{*}{-}                                                             & \multirow{2}{*}{2,952}                                                     & 9.61           & 2.64          & 9.14           & 16.13          \\ \cline{1-1} \cline{5-8} 
MobileNetV3              &                                                                                 &                                                                                &                                                                            & 10.42          & 4.54          & 10.90          & 21.36          \\ \hline
\textbf{EfficientNet-B2} & \multirow{2}{*}{\textbf{10\%}}                                                  & \multirow{2}{*}{\textbf{300}}                                                  & \multirow{2}{*}{\textbf{3,252}}                                            & \textbf{8.68}  & \textbf{3.47} & \textbf{7.06}  & \textbf{13.80} \\ \cline{1-1} \cline{5-8} 
\textbf{MobileNetV3}     &                                                                                 &                                                                                &                                                                            & \textbf{10.20} & \textbf{3.66} & \textbf{10.46} & \textbf{18.53} \\ \hline
EfficientNet-B2          & \multirow{2}{*}{20\%}                                                           & \multirow{2}{*}{600}                                                           & \multirow{2}{*}{3,552}                                                     & 10.09          & 3.15          & 10.08          & 19.79          \\ \cline{1-1} \cline{5-8} 
MobileNetV3              &                                                                                 &                                                                                &                                                                            & 13.47          & 7.50          & 18.27          & 29.30          \\ \hline
EfficientNet-B2          & \multirow{2}{*}{30\%}                                                           & \multirow{2}{*}{900}                                                           & \multirow{2}{*}{3,852}                                                     & 8.96           & 2.64          & 7.75           & 14.37          \\ \cline{1-1} \cline{5-8} 
MobileNetV3              &                                                                                 &                                                                                &                                                                            & 13.15          & 7.31          & 16.32          & 25.33          \\ \hline
EfficientNet-B2          & \multirow{2}{*}{50\%}                                                           & \multirow{2}{*}{1,500}                                                         & \multirow{2}{*}{4,452}                                                     & 10.21          & 4.03          & 10.46          & 18.95          \\ \cline{1-1} \cline{5-8} 
MobileNetV3              &                                                                                 &                                                                                &                                                                            & 12.22          & 6.18          & 14.49          & 25.21          \\ \hline
EfficientNet-B2          & \multirow{2}{*}{75\%}                                                           & \multirow{2}{*}{2,250}                                                         & \multirow{2}{*}{5,202}                                                     & 14.05          & 8.95          & 17.90          & 30.81          \\ \cline{1-1} \cline{5-8} 
MobileNetV3              &                                                                                 &                                                                                &                                                                            & 13.93          & 9.70          & 18.72          & 21.91          \\ \hline
EfficientNet-B2          & \multirow{2}{*}{100\%}                                                          & \multirow{2}{*}{5,904}                                                         & \multirow{2}{*}{5,904}                                                     & 11.62          & 5.80          & 13.49          & 22.50          \\ \cline{1-1} \cline{5-8} 
MobileNetV3              &                                                                                 &                                                                                &                                                                            & 10.97          & 4.91          & 11.78          & 21.11          \\ \hline
EfficientNet-B2          & \multirow{2}{*}{\begin{tabular}[c]{@{}l@{}}Only\\ synthetic\end{tabular}}       & \multirow{2}{*}{-}                                                             & \multirow{2}{*}{25,000}                                                    & 37.57          & 57.54         & 63.96          & 63.96          \\ \cline{1-1} \cline{5-8} 
MobileNetV3              &                                                                                 &                                                                                &                                                                            & 37.23          & 54.40         & 74.71          & 74.71          \\ \hline
\end{tabular}
    \end{threeparttable}
\end{table}

\section{Conclusions}
\label{sec:conclusions}
This study examines the impact of incorporating synthetic "non-morphed" data into the training process of a deep learning-based model for S-MAD. The findings indicate that solely relying on synthetic data for training results in suboptimal performance compared to the other scenarios. Additionally, incorporating a sample of synthetic data can enhance performance. However, the size of the sample added to the training process is contingent upon the specific dataset employed for deep learning-based model training.

Our study shows that compact architectures such as EfficientNet and MobileNet can provide strong baselines for morphing attack detection, while remaining suitable for realistic deployment scenarios. As a next step, we intend to extend our evaluation with additional backbone models to better understand the impact of model capacity, and to incorporate selected augmentation strategies that approximate operational conditions such as compression or mild print–scan artefacts. These directions will allow us to further assess the robustness of the proposed approach without departing from practical constraints.


\bibliographystyle{unsrt}
\bibliography{neurips_2025}

\end{document}